\documentclass[times， review， 10pt]{elsarticle}

\usepackage{amssymb}
\usepackage{float}
\usepackage{amsmath}
\usepackage{bm}
\usepackage{multirow}
\usepackage{makecell}
\usepackage{hyperref}
\usepackage{arydshln} 
\begin{document}

\begin{frontmatter}

%% Title, authors and addresses

%% use the tnoteref command within \title for footnotes;
%% use the tnotetext command for theassociated footnote;
%% use the fnref command within \author or \address for footnotes;
%% use the fntext command for theassociated footnote;
%% use the corref command within \author for corresponding author footnotes;
%% use the cortext command for theassociated footnote;
%% use the ead command for the email address,
%% and the form \ead[url] for the home page:
%% \title{Title\tnoteref{label1}}
%% \tnotetext[label1]{}
%% \author{Name\corref{cor1}\fnref{label2}}
%% \ead{email address}
%% \ead[url]{home page}
%% \fntext[label2]{}
%% \cortext[cor1]{}
%% \affiliation{organization={},
%%             addressline={},
%%             city={},
%%             postcode={},
%%             state={},
%%             country={}}
%% \fntext[label3]{}

\title{Dual-Attention Frequency Fusion at Multi-Scale for Joint Segmentation and Deformable Medical Image Registration}

%% use optional labels to link authors explicitly to addresses:
%% \author[label1,label2]{}
%% \affiliation[label1]{organization={},
%%             addressline={},
%%             city={},
%%             postcode={},
%%             state={},
%%             country={}}
%%
%% \affiliation[label2]{organization={},
%%             addressline={},
%%             city={},
%%             postcode={},
%%             state={},
%%             country={}}

%% or include affiliations in footnotes:
\author[phaddress1]{Hongchao Zhou}
\author[phaddress1]{Shunbo Hu \corref{cor1}}
\ead{hushunbo@lyu.edu.cn}

\cortext[cor1]{Corresponding author}
\affiliation[phaddress1]{organization={School of Information Science and Engineering},%Department and Organization
	addressline={Linyi University}, 
	city={Linyi},
	postcode={276000}, 
	state={Shandong},
	country={China}}
%\address[myeaddress]{These authors contributed equally to this manuscript}
%\date{}

\begin{abstract}
Deformable medical image registration is a crucial aspect of medical image analysis. 
In recent years, researchers have begun leveraging auxiliary tasks (such as supervised segmentation) to provide anatomical structure information for the primary registration task, addressing complex deformation challenges in medical image registration. 
In this work, we propose a multi-task learning framework based on multi-scale dual attention frequency fusion (DAFF-Net), which simultaneously achieves the segmentation masks and dense deformation fields in a single-step estimation.
DAFF-Net consists of a global encoder, a segmentation decoder, and a coarse-to-fine pyramid registration decoder. During the registration decoding process, we design the dual attention frequency feature fusion (DAFF) module to fuse registration and segmentation features at different scales, fully leveraging the correlation between the two tasks. The DAFF module optimizes the features through global and local weighting mechanisms. During local weighting, it incorporates both high-frequency and low-frequency information to further capture the features that are critical for the registration task. With the aid of segmentation, the registration learns more precise anatomical structure information, thereby enhancing the anatomical consistency of the warped images after registration. Additionally, due to the DAFF module's outstanding ability to extract effective feature information, we extend its application to unsupervised registration. Extensive experiments on three public 3D brain magnetic resonance imaging (MRI) datasets demonstrate that the proposed DAFF-Net and its unsupervised variant outperform state-of-the-art registration methods across several evaluation metrics, demonstrating the effectiveness of our approach in deformable medical image registration.
\end{abstract}


\end{frontmatter}

%% \linenumbers

%% main text
\section{Introduction}
Deformable image registration (DIR) \cite{ramadan2024medical} is a basic task in the field of computer vision, which is widely used in medical diagnosis, surgical guidance, disease detection, and so on \cite{haskins2020deep,zou2022review}, to determine an optimal spatial transformation that warps the moving image and align it with the corresponding fixed image. Traditional methods \cite{klein2009elastix} usually treat DIR as an optimization task, trying to minimize the energy function in an iterative manner, but it requires a lot of computation and long processing time \cite{klein2009evaluation}. Recently, deep learning (DL) methods have demonstrated exceptional performance in computer vision tasks, such as image classification \cite{kumar2024medical} and segmentation \cite{manakitsa2024review}. In DIR, deep learning (DL) methods are widely used due to their faster and more accurate registration than traditional methods. Hu et al.\cite{hu2017learning} demonstrate the ideal of learning registration is promising. DL methods use large datasets for training and employ an end-to-end approach to train network parameters. During the inference stage, the trained network can estimate the deformation field between input image pairs within a minimal time.

DL registration methods can be categorized into supervised, unsupervised, and weakly supervised approaches based on their training methods. Supervised methods typically require considerable supervised information, such as ground-truth DDFs \cite{sokooti2017nonrigid,wang2020deepflash}. However, obtaining ground-truth DDFs through expert clinicians is a challenging task, which limits the clinical feasibility of supervised methods. Unsupervised methods do not rely on any labeled data, they primarily align the moving image to the fixed image by calculating the similarity between image structures and applying smoothness constraints to the DDF. Current unsupervised methods, while capable of achieving fast registration, still exhibit limitations in registration accuracy and handling complex deformations. This issue arises partly from the inherent limitations of the registration network and partly from the lack of topological constraints on the regions of interest (ROIs).

To address the limitations of registration network performance, an effective solution is to improve the network architecture. With the outstanding performance of Transformer in various clinical applications \cite{he2023transformers}, such as image reconstruction, segmentation, and detection, researchers are beginning to explore their potential in DIR. TransMorph \cite{chen2022transmorph} designs a hybrid network structure combining Transformer and CNN, leveraging the self-attention mechanism and large receptive fields of Transformer to enhance registration performance. TransMatch \cite{chen2023transmatch} sends fixed and moving images through separate branches for feature extraction and uses a local window cross-attention module to perform explicit multilevel feature matching between the image pairs. However, due to the high computational cost of the self-attention mechanism in Transformer-based networks, particularly when handling 3D medical image registration, there is a significant demand for GPU resources. To address this high computational cost, Transformer-based networks in DIR typically output registration results at a lower resolution and then upsample them to full resolution. Another effective way to enhance network performance is by using CNNs with a feature pyramid strategy to obtain multi-scale features, enabling coarse-to-fine registration. NICE-Net \cite{meng2022non} employs a dual-stream encoder structure to extract features at different scales and performs coarse-to-fine registration within a single iteration of the network, significantly improving registration accuracy. RDP \cite{wang2024recursive} designs a purely convolutional pyramid network using a step-by-step recursion strategy, combining high-level semantic information to predict the DDFs at each step and achieve coarse-to-fine registration. 

To address the lack of topological constraints on the ROIs, segmentation labels can be used to provide auxiliary information. VoxelMorph \cite{balakrishnan2019voxelmorph} mentions that it can provide additional anatomical information for the registration task by calculating the loss function between the fixed segmentation labels and the warped moving segmentation labels. RClaNet \cite{wu2023rclanet} provides anatomical structure information for the registration task by utilizing segmentation labels and their boundaries. To ensure the plausibility of registration, VM$_{\text{LSDF}}$ \cite{yang2023deep} leverages the segmentation labels to synthesize their local-signed-distance fields, offering more realistic anatomical information for the registration task. However, these approaches only provide auxiliary information for the registration task derived from the segmentation task and do not consider the underlying relationship and mutual enhancement potentials between the two tasks. Increasing research is integrating registration and segmentation tasks, using segmentation to identify ROIs and registration to establish correspondences between different images of the same anatomical structures. Current weakly supervised methods often use loss fusion and feature fusion to achieve coordinated registration and segmentation. Therefore, finding an appropriate fusion strategy is crucial for the effectiveness of weakly supervised methods. AutoFuse \cite{meng2023autofuse} focuses on the issue of feature fusion in weakly supervised methods and has developed a data-driven feature fusion strategy, achieving superior registration performance compared to the aforementioned methods. 

Although existing improvements have enhanced registration performance to some extent, there is still room for further optimization. By reviewing and reflecting on various registration methods in recent years, we conclude that an efficient feature fusion module is crucial for enhancing network performance. Whether for unsupervised or weakly supervised methods, an effective feature fusion strategy can facilitate both efficient and accurate registration.

In this paper, we introduce a learning framework for joint registration and segmentation tasks based on dual attention frequency fusion at multi-scale (DAFF-Net). We focus on how to leverage the coupling and decoupling relationships between these tasks and achieve information exchange through both loss and feature fusion. DAFF-Net consists of a global encoder (Global Encoder), a segmentation-specific decoder (Segmentation Decoder), a registration-specific decoder (Coarse-to-fine Pyramid Registration Decoder), a flow estimator block (FEB), and a dual attention frequency fusion (DAFF) module, and designs to fully exploit the correlation between registration and segmentation tasks. The global encoder is used to couple the features of two tasks by shared parameters. Additionally, during feature fusion, given that registration is the primary task, we use the DAFF module to decouple the features from both tasks by global and local attention weighting, thereby capturing the most beneficial information for the registration task. Furthermore, recognizing the DAFF module's ability to efficiently handle fusion features, we extend DAFF-Net to unsupervised registration tasks. Our primary contributions can be summarized as follows:
\begin{itemize}
	\item We propose a novel joint registration and segmentation framework that constructs an improved pyramid decoder to achieve coarse-to-fine registration. To the best of our knowledge, the pyramid decoder is first used in the joint registration and segmentation framework. At the same time, we further enhance registration performance by leveraging the anatomical information from the ROIs provided by supervised segmentation to assist in registration, and by combining loss fusion with multi-scale feature fusion to capture the correlation between the two tasks.
	
	\item We design a dual attention frequency fusion module that applies attention to the fused features from both global and local perspectives while integrating high-frequency and low-frequency feature information. This approach fully explores the correlation between registration and segmentation tasks and maximizes the contribution of segmentation to the main registration task.
	
	\item We further extend DAFF-Net to unsupervised registration tasks and compare both DAFF-Net and its unsupervised variant with various state-of-the-art registration algorithms. Extensive quantitative and qualitative results demonstrate that DAFF-Net significantly outperforms other methods in brain MRI registration, achieving higher registration accuracy. Additionally, the unsupervised variant also demonstrates excellent registration performance and achieves plausible deformation.
\end{itemize}

\begin{figure}[!t]
	\centerline{\includegraphics[width=1.0\textwidth]{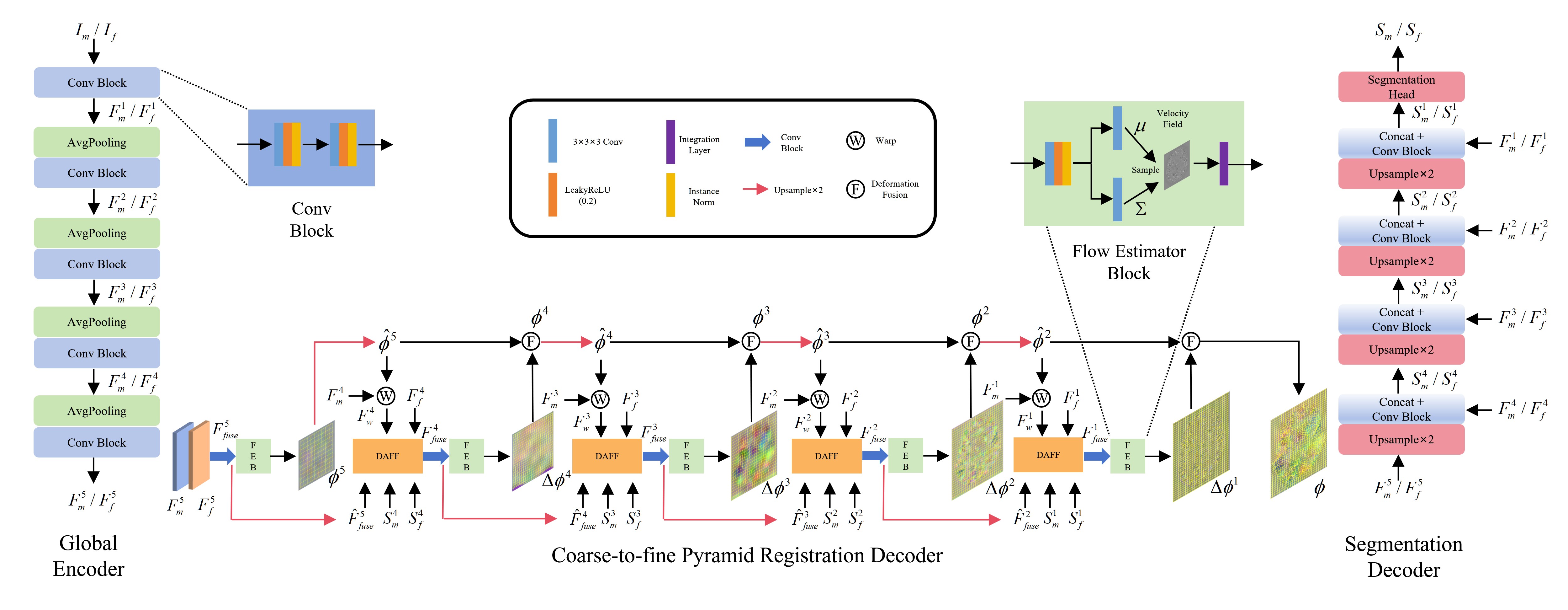}}
	\caption{The overview of DAFF-Net. DAFF-Net is a deep learning framework for joint registration and segmentation. It consists of a shared-parameter global encoder for two tasks, a pyramid-based registration decoder that operates from coarse to fine, and a shared-parameter decoder for segmentation tasks. The global encoder processes the moving ($I_m$) and fixed ($I_f$) images to extract hierarchical features, then the segmentation decoder uses these features to produce segmentation maps $S_m$ and $S_f$. The pyramid registration decoder includes the flow estimator block (FEB) and the dual attention frequency fusion (DAFF) module, which predict DDFs layer by layer and progressively combine them to obtain the final DDF $\phi$.}
	\label{fig1}
\end{figure}
\section{Related work}
\subsection{Pyramid-based Registration Method}
The pyramid strategy is widely used in both traditional \cite{klein2009evaluation,burt1987laplacian} and DL methods.
In DL methods, pyramid-based registration methods predict residual DDFs at different scales in the decoder and progressively combine these residuals to construct the final DDF, achieving precise coarse-to-fine registration. By reviewing recent pyramid-based registration methods, we further classify them based on their training approach into three categories: layered training pyramids, simultaneous training pyramids, and iterative loop pyramids.

LapIRN \cite{mok2020large} introduces a 3-level deep Laplacian pyramid network, which uses three identical CNN-based registration networks to optimize at different resolutions, achieving registration from coarse to fine. As the pyramid-based registration method with simultaneous training, Im2Grid \cite{liu2022coordinate} introduces multiple coordinate translators in the pyramid network and applies cross-correlation mechanisms at each layer of the decoder. NICE-Net \cite{meng2022non} optimizes the encoder and decoder structures, allowing the network to selectively propagate features as needed, and to progressively learn coarse-to-fine transformations in each iteration. ModeT \cite{wang2023modet} introduces the motion decomposition Transformer in the pyramid network and designs a competitive weighting module to merge multiple deformation sub-fields, generating the final DDF. As the pyramid-based registration methods with iterative loop, PIViT \cite{ma2023pivit} incorporates the Swin Transformer-based decoder in the network, performing iterative registration at low-scale to capture the coarse distribution of DDFs, followed by high-scale feature refinement to achieve a highly accurate DDF. Both RDP \cite{wang2024recursive} and IIRP-Net \cite{ma2024iirp} use iterative strategies at all scales. RDP employs a step-by-step recursive approach during training, integrating high-level semantic information to predict the DDF. IIRP-Net, on the other hand, applies an iterative inference strategy during the inference stage, fully leveraging the registration capability of the proposed flow estimator.

\subsection{The Framework for Joint Registration and Segmentation}

In the field of DIR, the goal of joint registration and segmentation frameworks is to enhance the performance of both tasks by optimizing two networks simultaneously, leveraging their correlation. After reviewing existing approaches, joint registration and segmentation frameworks can be classified into three categories: loss fusion, feature fusion, and parameter-sharing. Loss fusion optimizes the combined loss functions of registration and segmentation, with typical examples including DeepAtlas \cite{xu2019deepatlas}, DeepRS \cite{he2020deep}, and RSegNet \cite{qiu2021rsegnet}. DeepAtlas introduces the first deep neural network framework for joint learning of registration and segmentation tasks. This approach enables both tasks to learn from the same set of images during training, ultimately improving registration performance through anatomy similarity loss. DeepRS enhances registration by introducing random perturbation factors for sustainable data augmentation. An alignment confidence map method and a deep-based region constraint strategy are proposed to fully leverage the supervisory information from the segmentation task, enabling precise registration in ROIs. RSegNet is a joint learning framework that integrates a diffeomorphic registration network with a segmentation network. It introduces consistency supervision in diffeomorphic registration, leveraging auxiliary information from segmentation to achieve more accurate deformations. Both feature fusion and parameter-sharing approaches incorporate loss fusion. In parameter sharing methods, The parameter-sharing methods facilitate collaboration between tasks by sharing parameters in certain convolutional layers, leveraging the shared features to improve overall task performance. U-ReSNet \cite{estienne2019u} and JRP-Net \cite{zhao2021deep} use a shared encoder to learn common features for two tasks, and then employ two task-specific decoders to perform each task separately. U-ReSNet integrates affine registration within the network to generate precise transformation maps, while JRP-Net combines registration and parcellation tasks in a joint learning framework, utilizing parcellation map similarity loss to enhance boundary consistency and provide auxiliary information for registration. Cross-stitch \cite{elmahdy2021joint} employs a cross-stitch network architecture to facilitate information flow between the registration and segmentation tasks, while also exploring how the two tasks interact through different loss weighting strategies. In feature fusion, the method enhances task collaboration by integrating the feature information from both tasks during the registration network. AC-DMiR \cite{khor2023anatomically} proposes an anatomically aware DIR method, designing a cross-task attention block to capture the correlation between the two tasks by combining their initially predicted bottleneck features. In the final registration, AC-DMiR introduces voxel-wise weight to assign higher weights to regions with weaker anatomical relevance, thereby refining the final DDF. AutoFuse \cite{meng2023autofuse} designs a fusion gate module to combine the feature information from both tasks at different positions and proposes a data-driven fusion strategy. It is the first deep registration method to employ a fusion strategy independent of empirical design. BFM-Net \cite{zhang2024bi} is a joint registration and segmentation framework based on the bi-fusion of structure and deformation at multi-scale. It effectively improves registration performance by transferring latent feature representations between registration and segmentation tasks through a multi-task connection module. 

While the aforementioned methods explore ways to improve registration accuracy by enhancing network architecture or integrating registration and segmentation frameworks, to the best of our knowledge, no method has combined both approaches. Existing multi-task frameworks typically focus on task correlation but often overlook the direct impact of the registration network architecture on performance. Our proposed DAFF-Net addresses both aspects: it leverages loss fusion, feature fusion, and parameter sharing to fully explore the correlation between registration and segmentation tasks. Additionally, by incorporating a pyramid strategy into the registration network, we enhance registration accuracy through feature fusion on different scales.
\section{Method}
\subsection{Network Architecture}
Figure \ref{fig1} illustrates the proposed DAFF-Net architecture, which consists of a global encoder, a coarse-to-fine pyramid registration decoder, and a segmentation decoder. The registration decoder includes a flow estimator block (FEB) and a dual attention frequency fusion (DAFF) module. It efficiently integrates registration and segmentation features across different scales, calculates residual DDFs layer by layer, and progressively fuses them to generate the final DDF $\phi$.
\subsubsection{Global Encoder}
 Due to the coupling between the registration and segmentation tasks, we design a global encoder to learn common features for both. We select a powerful and efficient conv block (ConvB) as the feature extraction backbone for the encoder. Each ConvB consists of two conv sub-blocks. Each conv sub-block includes a 3$\times$3$\times $3 convolutional layers, a LeakyReLU activation function (with parameter 0.2), and instance normalization. 
$I_m$ and $I_f$ are separately input into the shared-parameter global encoder, producing feature maps $F_{m}^{1} $ and $F_{f}^{1} $ through the first ConvB. In the following four layers, the feature maps are first downsampled using average pooling, with each layer followed by a ConvB. Thus, the global encoder produces two sets of feature maps,  $F_{m}^{i} $ and  $F_{f}^{i} $, where $i$ ranges from 1 to 5. The channel numbers of the global encoder are set of [16, 32, 32, 64, 64].

\subsubsection{Segmentation Decoder}
For the segmentation task, we design a task-specific decoder (segmentation decoder). During the decoding process, two decoders with shared parameters process the feature maps $F_{m}^{i} $ and $F_{f}^{i} $ generated by the global encoder. For example, when processing the lowest scale, the feature map $F_{m}^{5} $ is first upsampled. The upsampled feature map is then combined with the low-level feature map $F_{m}^{4} $ of the same scale and is fed into ConvB, generating the segmentation feature map $S_{m}^{4} $ for that layer. The operations for the remaining three layers are similar to those of the first layer.
Thus we sequentially generate segmentation feature maps $S_{m}^{5-j} $ and $S_{f}^{5-j} $ through the four layer structure of the decoder, where $j$ ranges from 1 to 4. Finally, $S_{m}^{1} $ and $S_{f}^{1} $ are fed into a convolutional layer with a softmax activation (segmentation head) to produce the final segmentation masks $S_{m}$ and $S_{f}$. The channel numbers of the segmentation decoder are set of [64, 32, 32, 16, 4].
\begin{figure}[!t]
	\centerline{\includegraphics[width=1.0\textwidth]{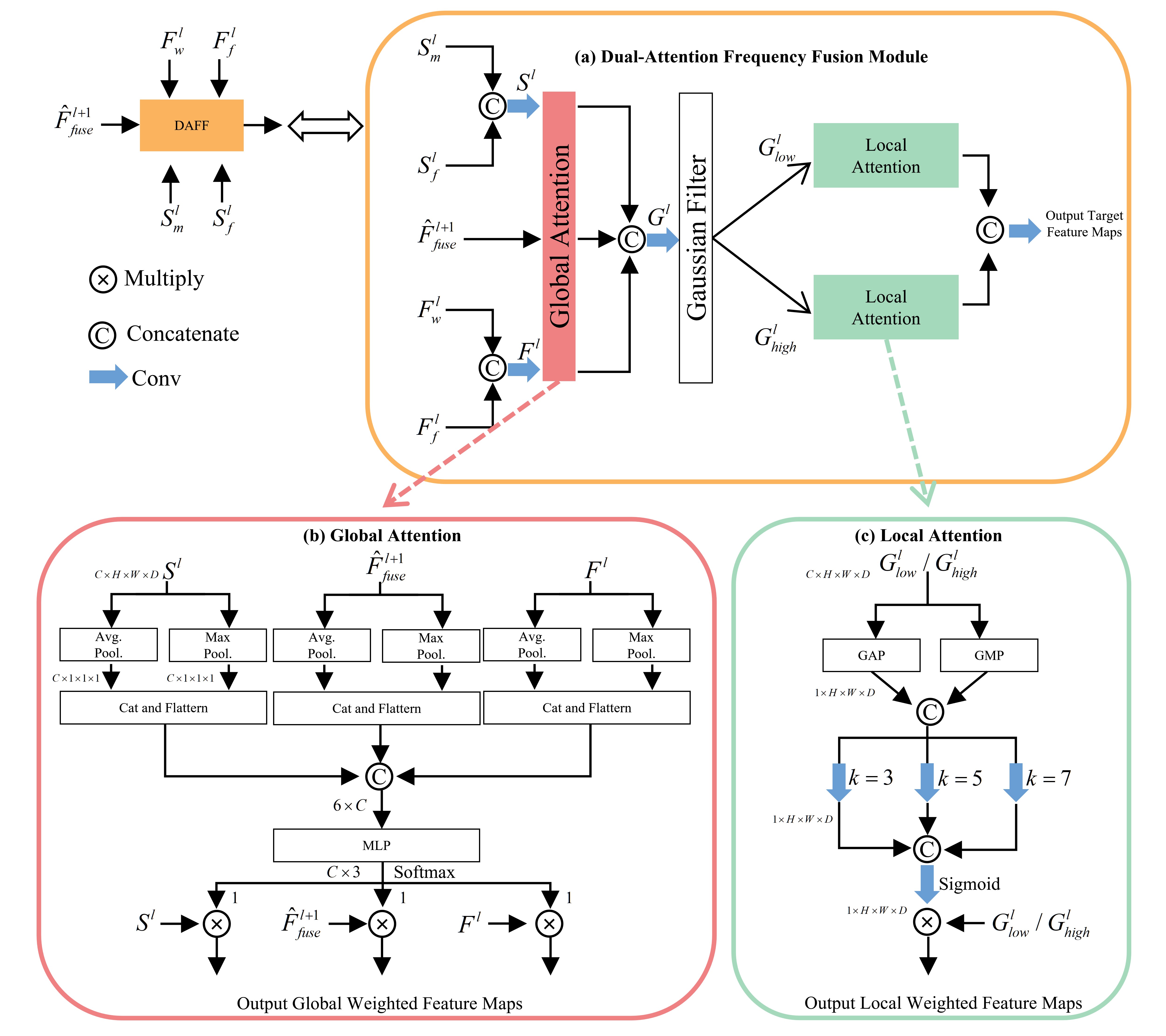}}
	\caption{The proposed Dual-Attention Frequency Fusion (DAFF) Module.}
	\label{fig2}
\end{figure}

\subsubsection{Coarse-to-fine Pyramid Registration Decoder} The coarse-to-fine pyramid registration decoder is a task-specific decoder designed for registration, which is the main focus of this paper. Its primary function is to fuse feature maps $F_{m}^{i} $ and $F_{f}^{i} $ from the global encoder with feature maps $S_{m}^{5-j}$ and $S_{f}^{5-j}$ from the segmentation decoder. It predicts sub-fields $\phi^{l}$ layer by layer across five scales ($l$ = 5, 4, 3, 2, 1), and gradually combines these sub-fields to generate the final precise DDF $\phi$. The core components of the registration decoder include the flow estimator block (FEB) and the dual attention frequency fusion (DAFF) module.
The FEB is designed to calculate the velocity field and DDF. Inspired by existing methods \cite{meng2023autofuse,dalca2019unsupervised} for velocity field computation, we fed the output of the ConvB into a conv sub-block of FEB to increase the network depth, and then the resulting feature map is fed into two parallel convolutional layers to obtain the mean ($\mu$) and variance ($\Sigma$) of the velocity field. The velocity field is generated by sampling from $\mu$ and $\Sigma$, which is then converted into a DDF through a differentiable squaring and scaling integration layer. The DAFF module is designed to deal with the feature fusion problem in multi-task learning, which can efficiently capture features most beneficial for the registration task (detailed in Section \ref{section 3.2}).

At the coarsest scale ($l$=5), the concatenated features of $F_{m}^{5}$ and  $F_{f}^{5} $ are processed through ConvB to produce the fused feature map ($F_{fuse}^{5}$). $F_{fuse}^{5}$ is then fed into the FEB to generate the DDF $\phi^{5}$. The procedure can be described as follows:
\begin{equation}
	\left\{
	\begin{aligned}
		F_{fuse}^{l}&=ConvB(F_{m}^{l},F_{f}^{l}),\\
		\phi^{l}&=MFB(F_{fuse}^{l}),
	\end{aligned} \right.
	l=5.
	\label{eq1}
\end{equation}

At each subsequent scale $l$, the fused feature maps $F_{fuse}^{l+1}$ are first upsampled to generate $\hat{F}_{fuse}^{l+1}$, and the DDF $\phi^{l+1}$ is upsampled to generate $\hat{\phi}^{l+1}$. $\hat{\phi}^{l+1}$ is then used to warp $F_{m}^{l}$, resulting in the warped moving feature map $F_{w}^{l}$. Next, $F_{w}^{l}$, $F_{f}^{l}$, $\hat{F}_{fuse}^{l+1}$, $S_{m}^{l}$, and $S_{f}^{l}$ are all fed into the DAFF module. The output of the DAFF module is then fed into ConvB to generate $F_{fuse}^{l}$, and $F_{fuse}^{l}$ is then fed into FEB to acquire the residual DDF $\triangle \phi^{l}$. The composition of $\hat{\phi}^{l+1}$ and $\triangle \phi^{l}$ constitutes the deformation sub-field $\phi^{l}$ at the current scale. The procedure at the subsequent scale can be described as follows:
\begin{equation}
	\left\{
	\begin{aligned}
		\hat{F}_{fuse}^{l+1}&=up(F_{fuse}^{l+1}),\\
		\hat{\phi}^{l+1}&=up(\phi^{l+1}),\\
		F_{w}^{l}&=F_{m}^{l}\circ \hat{\phi}^{l+1},\\
		F_{fuse}^{l}&=ConB(DAFF(F_{w}^{l},F_{f}^{l},\hat{F}_{fuse}^{l+1},S_{m}^{l},S_{f}^{l})),\\
		\triangle \phi^{l}&=FEB(F_{fuse}^{l}),\\
		\phi^{l}&=\hat{\phi}^{l+1} \circ \triangle \phi^{l},
	\end{aligned} \right.
	l\in [4,3,2,1],
	\label{eq2} 
\end{equation}
where $up$ denotes to upsampling using trilinear interpolation and $\circ$ denotes composite transform. $\phi^{1}$ is the final DDF $\phi$.

\subsection{Dual-attention Frequency Fusion Module} \label{section 3.2}
To fully leverage the correlation between registration and segmentation and to enhance segmentation's contribution to registration, we fed the feature maps generated by the global encoder and segmentation decoder into the cross-task feature fusion module. This process is achieved through our proposed dual attention frequency fusion (DAFF) module. The DAFF module captures the interrelationship between registration and segmentation by integrating the coupled encoded feature maps, task-specific decoded feature maps, and the registration fused feature maps from the previous scale. Unlike directly concatenating all features, the DAFF module captures the most beneficial features for registration by decoupling the features of different tasks. The proposed DAFF module structure is depicted in Fig\ref{fig2}.

In the coarse-to-fine pyramid registration decoder, the DAFF module is utilized at all scales except for the coarsest one. DAFF module consists of two main components: global attention and local attention. At scale $l$=4, for instance, $S_{m}^{l}$ and $S_{f}^{l}$ are first concatenated, followed by a convolution operation to obtain the segmentation feature map $S^{l}$. Similarly, $F_{w}^{l}$ and $F_{f}^{l}$ are concatenated, and a convolution operation is applied to generate the coupled feature map $F^{l}$. After applying global attention to $S^{l}$, $F^{l}$, and $\hat{F}_{fuse}^{l+1}$, we obtain the global weighted fused feature maps $G^{l}$ through concatenation and convolution operations. We then process $G^{l}$ using the Gaussian Filter to generate low-frequency feature maps $G_{low}^{l}$ and high-frequency feature maps $G_{high}^{l}$. The smoothing effect of the low-frequency Gaussian Filter removes noise from the original feature maps, so the low-frequency feature maps capture the overall shape and main anatomical structures of the image. The high-frequency information retains detailed features, particularly edges and boundaries between tissues, which are crucial for capturing complex local deformations in registration tasks. Finally, $G_{low}^{l}$ and $G_{high}^{l}$ are each fed into the local attention module, and the resulting features are concatenated and convolved to produce the fused features that optimally correspond to the registration task. The procedure of the DAFF module can be described as follows:
\begin{equation}
	\left\{
	\begin{aligned}
		S^{l}&=CConv(S_{m}^{l},S_{f}^{l}), \\
		F^{l}&=CConv(F_{w}^{l},F_{f}^{l}), \\
		G^{l}&=CConv(GlobalAtt(S^{l},F^{l},\hat{F}_{fuse}^{l+1})),\\
		G_{low}^{l}&=GaussianFilter(G^{l}),\\
		G_{high}^{l}&=G^{l}-G_{low}^{l},\\
		output&=CConv(LocalAtt(G_{low}^{l}),LocalAtt(G_{high}^{l})),
	\end{aligned} \right.
	l \in [4,3,2,1],
	\label{eq3} 
\end{equation}
where $CConv$ denotes the concatenate and convolution operation.

\subsubsection{Global Attention} 
The global attention is designed to assign weights to $S^{l}$, $F^{l}$, and $\hat{F}_{fuse}^{l+1}$ before their concatenation, the architecture is shown in Fig \ref{fig2} (b). Although registration and segmentation tasks are coupled, not all feature information is beneficial for registration. Therefore, we design the global attention to weight the three feature maps across the global range. In the global attention process, for $S^{l}$, $F^{l}$, and $\hat{F}_{fuse}^{l+1}$, adaptive average pooling (Avg. Pool.) and adaptive max pooling (Max. Pool.) are first applied separately. The results are then flattened and concatenated and are fed into a Multilayer Perceptron (MLP) to obtain a projected feature map. Finally, the weights of the three feature maps are computed using a Softmax function, and $S^{l}$, $F^{l}$, and $\hat{F}_{fuse}^{l+1}$ are weighted accordingly by these weights.

\subsubsection{Local Attention} 
The architecture of the local attention, as shown in Fig \ref{fig2} (c), is inspired by the content-guided attention proposed by Chen et al. \cite{10411857}. For both $G_{low}^{l}$ and $G_{high}^{l}$, global average pooling (GAP) and global max pooling (GMP) are applied across the channel dimension, and the results are concatenated. Next, convolutions with kernels of different sizes ($k$ = 3, 5, 7) are applied to capture information at various scales. Smaller kernels capture fine details and local features, while larger kernels expand the receptive field to capture broader contextual information. By combining the convolutions of these kernels of different sizes, the network can extract richer features across multiple scales, enhancing its ability to recognize complex structures. The features processed by the three convolutionals are then concatenated and passed through a convolutional layer with sigmoid activation to produce a spatial importance map, which is finally used to weight $G_{low}^{l}$ and $G_{high}^{l}$.

With such a design, we can fully explore the latent connections between registration and segmentation tasks and effectively leverage segmentation to enhance registration. The global attention weights and decouples feature maps from different sources, focusing on features that are beneficial to the registration task during concatenation. The local attention, by computing spatial importance maps, assigns weights to the high-frequency and low-frequency feature maps separately. Using convolutional kernels of different sizes, local attention captures multi-scale information, enhancing the model's ability to handle complex structures. By applying local attention from both high- and low-frequency perspectives, we preserve the overall structural information of the image while capturing fine local details. As a result, we learn the correlation between registration and segmentation tasks, ensuring that registration fully benefits from the anatomical information provided by segmentation.

\begin{figure}[!t]
	\centerline{\includegraphics[width=1.0\textwidth]{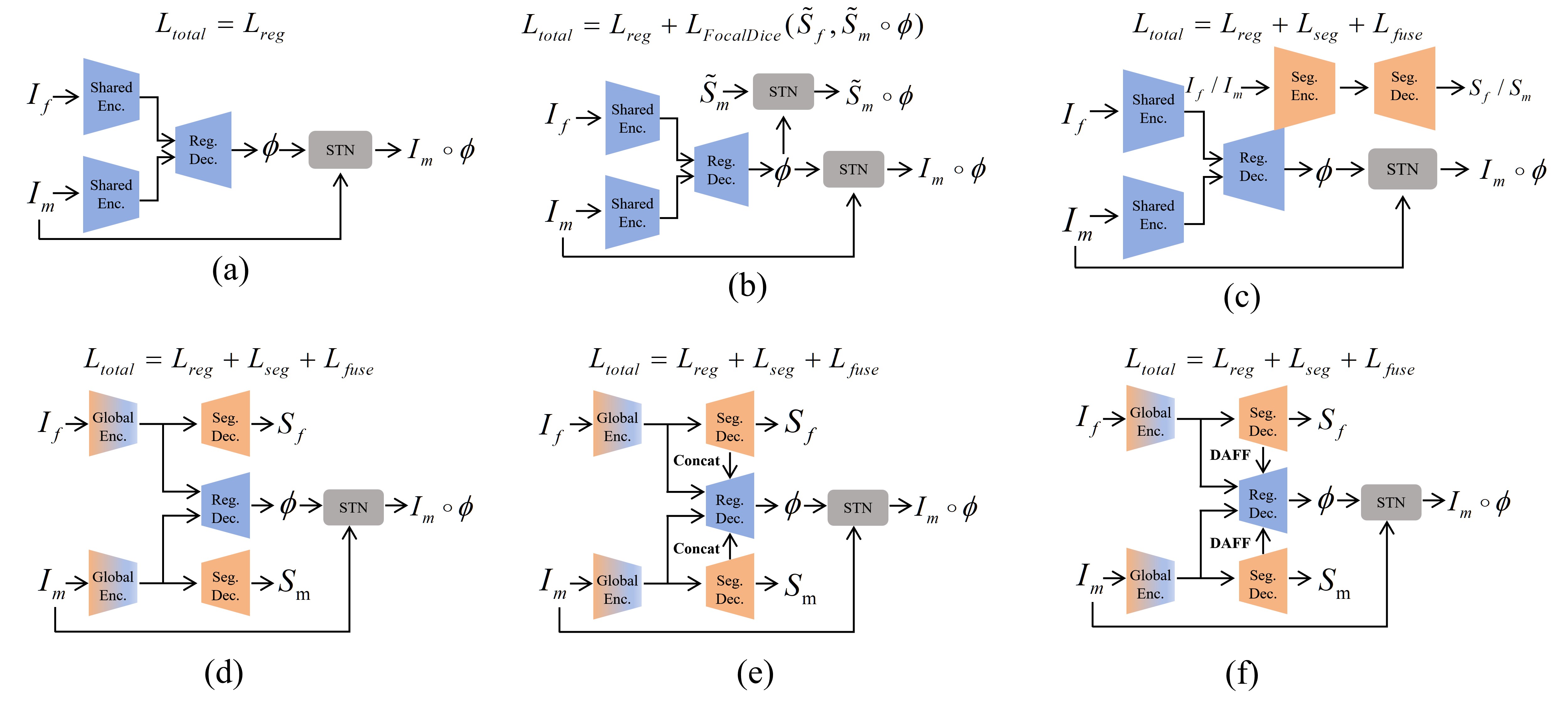}}
	\caption{The network architectures used in the ablation study. (a) PyramidReg is the base pyramid registration network; (b) AuxReg builds on PyramidReg by calculating a fusion loss using segmentation labels, providing weakly supervised information for registration; (c) SimSReg combines PyramidReg with an auxiliary U-Net segmentation network; (d) GloSReg is a coupled architecture comprising the shared global encoder and task-specific decoders; (e) CcSReg integrates registration and segmentation features through a simple concatenation operation; (f) represents the complete DAFF-Net architecture. The objective functions used for training each ablation network are illustrated in the figure, with detailed descriptions provided in Section \ref{section 5.2}.}
	\label{fig3}
\end{figure}

\subsection{Loss Function}
The objective function $L_{total}$ of DAFF-Net consists of three components: registration loss $L_{reg}$, segmentation loss $L_{seg}$, and a fusion term $L_{fuse}$ that combines registration and segmentation.

The registration loss $L_{reg}$ is composed of similarity loss $L_{sim}$, L2 regularization $L_{smooth}$, and the negative Jacobian determinant (NJD) $L_{njd}$. DAFF-Net uses the normalized cross correlation (NCC) \cite{rao2014application} to evaluate the similarity between the fixed image $I_f$ and the warped moving image $I_{w}=I_{m} \circ \phi$. The similarity loss $L_{sim}$ can be defined as:
\begin{equation}
	L_{sim}(I_{f},I_{m}\circ \phi)=-\sum_{p\in \Omega}\frac{ {\textstyle (\sum_{p_i}(I_{f}(p_i)-\bar{I_f}(p))(I_w(p_i)-\bar{I_w}(p))})^{2}}{( {\textstyle \sum_{p_i}(I_f(p_i)-\bar{I_f}(p))^{2})( {\textstyle \sum_{p_i}(I_w(p_i)-\bar{I_w}(p))^{2})}}}  , \label{eq4}   
\end{equation}
where $\Omega$ represents the whole volumetric domain, $\bar{I_f(p)}$ and $\bar{I_w(p)}$ represent the average local voxel value, $p_i$ represents the local neighborhood in $n^{3}$ volumetric patch centered at $p$, and $n$ is set to 9 in our experiments.

To ensure the smoothness of the DDF $\phi$, we use L2 regularization as a deformation smooth regularizer:
\begin{equation}
L_{smooth}(\phi)=\sum_{p\in \Omega}\left \| \bigtriangledown \phi(p) \right \|^{2}.    \label{eq5}   
\end{equation}

In DAFF-Net, we also use an additional regularization term to penalize deformations:
\begin{equation}
	L_{njd}(\phi)=\sum_{p\in \Omega}0.5(\left | det(\bigtriangledown \phi(p)) \right |-det(\bigtriangledown \phi (p)) ) .    \label{eq6}   
\end{equation}

Therefore, the registration loss $L_{reg}$ is:
\begin{equation}
	L_{reg}=L_{sim}(I_{f},I_{m}\circ \phi)+L_{smooth}(\phi)+\lambda _{0}L_{njd}(\phi), \label{eq7}
\end{equation}
where $\lambda _{0}$ is hyperparameter. To explore the generalizability of the proposed DAFF module in unsupervised network, we design an unsupervised variant of DAFF-Net (DAFF-Net (uns)). The architecture of this variant is identical to the original, except for the removal of the segmentation head from the auxiliary decoder (as shown in Figure \ref{fig1}). Its objective function only includes the registration loss $L_{reg}$.

To calculate the segmentation loss $L_{seg}$, we use Dice loss \cite{milletari2016v} $L_{Dice}$ and Focal loss \cite{lin2017focal} $L_{Focal}$ to measure the overlap between the segmentation masks $S$ (consists of $S_m$ and $S_f$) predicted by the network and the ground truth segmentation labels $\widetilde{S}$ (consists of the moving and fixed segmentation labels $\widetilde{S}_m$ and $\widetilde{S}_f$). Before calculating $L_{Dice}$ and $L_{Focal}$, we first convert the input data to one-hot encoding. The expression for the $L_{Dice}$ is as follows:
\begin{equation}
	L_{Dice}(\widetilde{S},S)=-\frac{1}{K}\sum_{k=1}^{K}\frac{2 {\textstyle \sum_{p\in \Omega}\widetilde{S}_{k}(p)\cdot S_{k}(p)}}{ {\textstyle \sum_{p\in\Omega}\widetilde{S}_{k}(p)+ {\textstyle \sum_{p\in \Omega}S_{k}(p)}  } }  , \label{eq8} 
\end{equation}
where $K$ denotes the number of brain tissue in the segmentation labels. 

To address the potential extreme imbalance between foreground and background classes during segmentation training, we use the Focal loss $L_{Focal}$ to optimize the segmentation. Focal loss introduces a modulation factor $\gamma$ and a weighting factor $\alpha$ to the traditional cross-entropy loss, to improve the model's focus on hard-to-classify positive samples and minority class samples. In this study, we adopted the configuration of $\gamma$=2 and $\alpha$=0.25, based on the optimal settings reported by Lin et al. \cite{lin2017focal} in their experiments. Therefore, the segmentation loss $L_{seg}$ is:
\begin{equation}
	\begin{cases}L_{FocalDice}=L_{Focal}+L_{Dice},
		\\L_{seg}=L_{FocalDice}(\widetilde{S}_{m},S_{m})+L_{FocalDice}(\widetilde{S}_{f},S_{f}).
	\end{cases}
	\label{eq9}
\end{equation}

The fusion term $L_{fuse}$ introduces mutual constraints between registration and segmentation, further exploring the correlation between the two through the integration of their loss functions. The $L_{fuse}$ is defined as follows:
\begin{equation}
	\begin{split}
		L_{fuse}=&L_{FocalDice}(\widetilde{S}_{f},\widetilde{S}_{m}\circ \phi)+L_{FocalDice}(\widetilde{S}_{f},S_{m}\circ \phi)\\
		&+L_{FocalDice}(S_{f},S_{m}\circ \phi).
	\end{split}
	\label{eq10}
\end{equation}

By combining the aforementioned loss functions, the overall objective function $L_{total}$ for DAFF-Net can be defined as:
\begin{equation}
	\begin{split}
		L_{total}&=L_{reg}+\lambda _{1}L_{seg}+\lambda _{2}L_{fuse}\\
		&=L_{sim}(I_f,Im\circ \phi)+L_{smooth}(\phi)+\lambda _{0}L_{njd}(\phi)\\
		&+\lambda _{1}( L_{FocalDice}(\widetilde{S}_m,Sm)+ L_{FocalDice}(\widetilde{S}_f,Sf) )\\
		&+\lambda _{2}(L_{FocalDice}(\widetilde{S}_f, \widetilde{S}_{m}\circ \phi )+L_{FocalDice}(\widetilde{S}_f,S_{m}\circ \phi)\\
		&+ L_{FocalDice}(S_f,S_{m}\circ \phi )),
	\end{split}
	\label{eq11} 
\end{equation}
where $\lambda_{0}$, $\lambda_{1}$ and $\lambda_{2}$ denote the weights of the regularization term $L_{njd}$, segmentation loss $L_{seg}$ and fusion term $L_{fuse}$, respectively. For parameter experiments refer to Section \ref{section 5.3}.
\section{Experiments}
\begin{figure}[!t]
	\centerline{\includegraphics[width=1.0\textwidth]{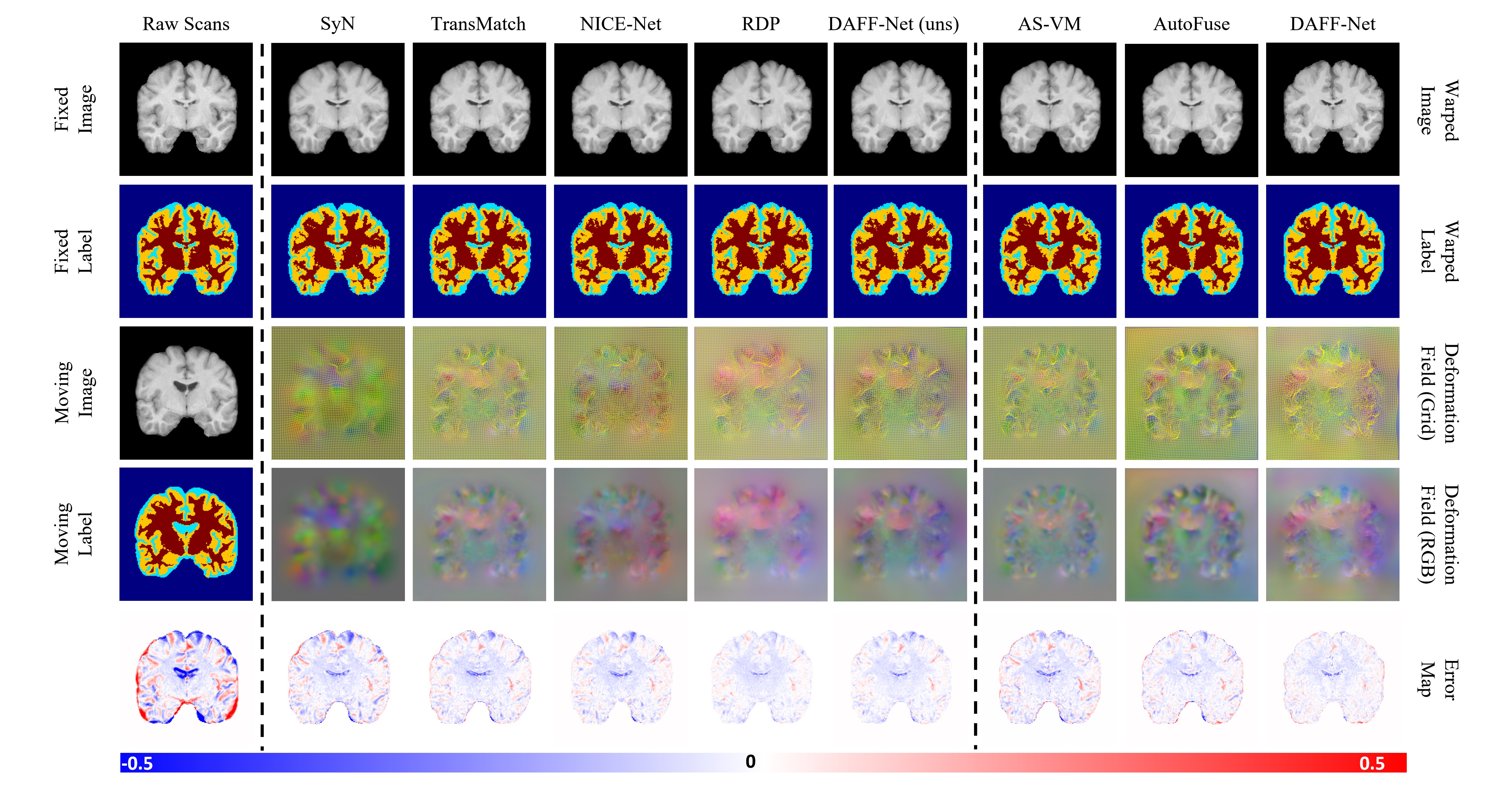}}
	\caption{Visual results of different registration methods on the LPBA40 dataset. The first column includes the fixed image, fixed image segmentation label, moving image, moving image segmentation label, and voxel-wise difference (error map) between the moving and fixed images. From top to bottom, excluding the first column: warped moving image, warped moving image segmentation labels, estimated deformations (shown in grid), estimated deformations (shown in RGB), and error maps between the warped moving image and the fixed image.}
	\label{fig4}
\end{figure}
\subsection{Datasets and  Preprocessing}
\subsubsection{LPBA40}
The 3D brain MRI dataset LPBA40 \cite{shattuck2008construction} includes 40 brain MRI images with annotated segmentations for cerebrospinal fluid (CSF), gray matter (GM), and white matter (WM). All images are preprocessed by affine registration to MNI space using FSL \cite{jenkinson2012fsl} tools, followed by min-max normalization and cropping to a volume size of 160$\times$192$\times $160 $mm^{3}$ with a voxel size of 1$\times$1$\times1$ $mm^{3}$. We randomly select 30 images for training and 10 images for testing. During training, two images are randomly chosen in each iteration, designated as the moving and fixed images. For testing, each image is evaluated as a fixed image, resulting in a total of 90 image pairs.
\subsubsection{OASIS1}
The 3D brain MRI dataset OASIS1 \cite{marcus2007open} consists of 416 brain MRI images, with a data processing pipeline identical to that of the LPBA40 dataset. We randomly select 366 images for training and 50 images for testing. For testing, we generate a total of 2450 image pairs.
\subsubsection{OASIS3}
The 3D brain MRI dataset OASIS3 \cite{lamontagne2019oasis} consists of 1742 brain MRI images, with a data processing pipeline identical to that of the LPBA40 dataset. We randomly select 1442 images for training and 50 images for testing. For testing, we generate a total of 2450 image pairs. 

\begin{figure}[!t]
	\centerline{\includegraphics[width=1.0\textwidth]{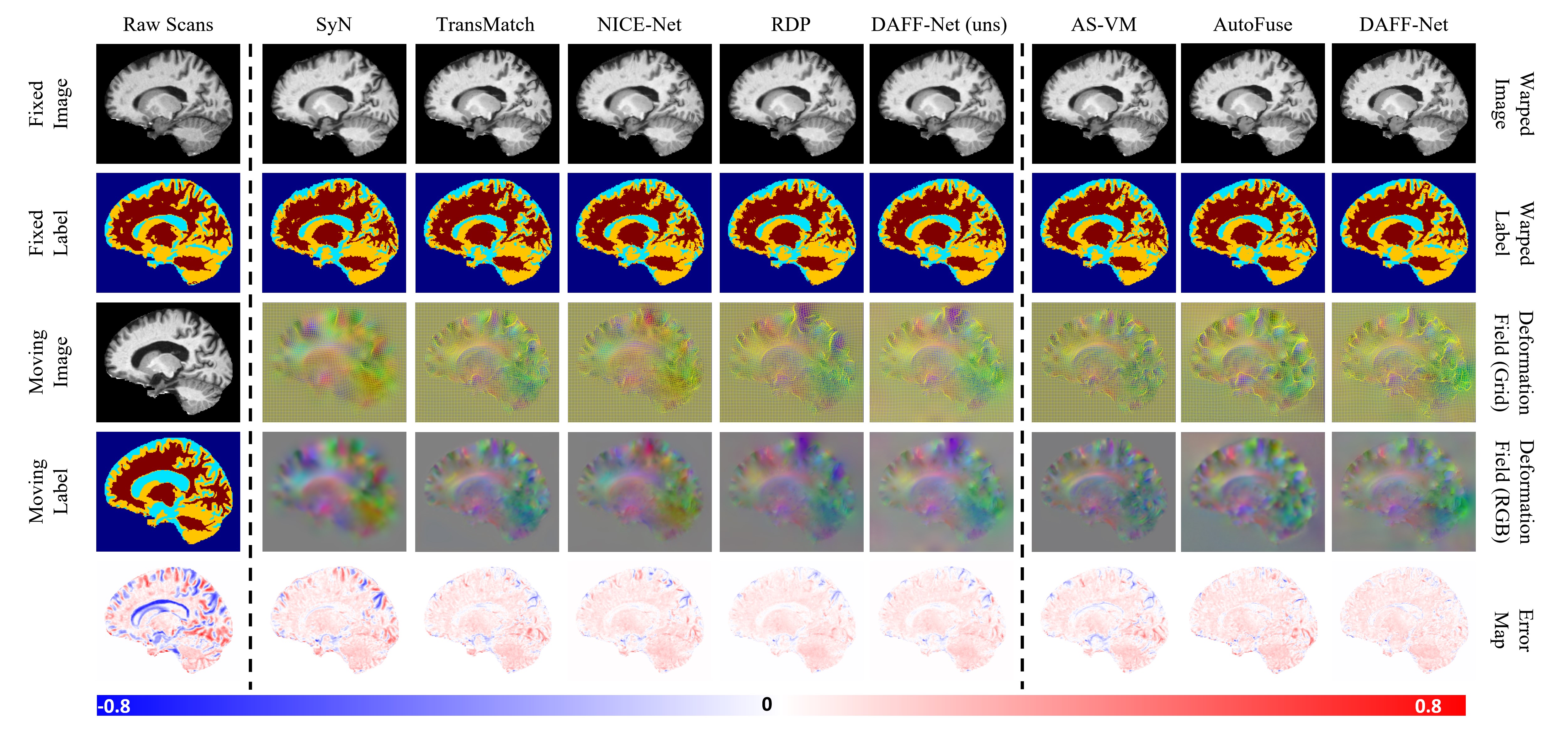}}
	\caption{Visual results of different registration methods on the OASIS1 dataset.}
	\label{fig5}
\end{figure}

\subsection{Comparison Methods}
We compare DAFF-Net and its unsupervised variant with several state-of-the-art registration methods: (1) SyN \cite{avants2008symmetric}: a classical traditional registration method, we employ the Cross Correlation (CC) as the objective function, and the number of iterations at three different scales is set to [10, 10, 5]. (2) TransMatch \cite{chen2023transmatch}: a dual-stream Transformer-based network. (3) NICE-Net \cite{meng2022non}: a non-iterative pyramid network with coarse-to-fine registration. (4) RDP \cite{wang2024recursive}: a novel pyramid registration network using a step-by-step recursion strategy. (5) AS-VM \cite{balakrishnan2019voxelmorph}: a classical deep learning network, that uses segmentation labels to provide auxiliary supervised segmentation information for registration. (6) AutoFuse \cite{meng2023autofuse}: a state-of-the-art deep learning framework that jointly performs image registration and image segmentation. For all comparison methods, we use the optimal parameter settings as reported in their respective papers. For NICE-Net and AutoFuse, which use the $L_{njd}$, we set the weight of this loss to be equal to $\lambda_{0}$.

\subsection{Implementation and training details} \label{section 4.3}
Our experiments are implemented using PyTorch on NVIDIA Tesla V100 with 32GB memory. We employ the Adam optimizer with a learning rate of 0.0001, and the batch size is set to 1. During the training phase, the LPBA40 dataset is trained for 800 epochs, the OASIS1 dataset for 300 epochs, and the OASIS3 dataset for 150 epochs. The hyperparameters in Eq. \ref{eq11} are set as follows: $\lambda_{0} = 1\times10^{-5}$ \cite{meng2023autofuse}, $\lambda_{1} = 0.5$ and $\lambda_{2} = 1$. 

\begin{figure}[!t]
	\centerline{\includegraphics[width=1.0\textwidth]{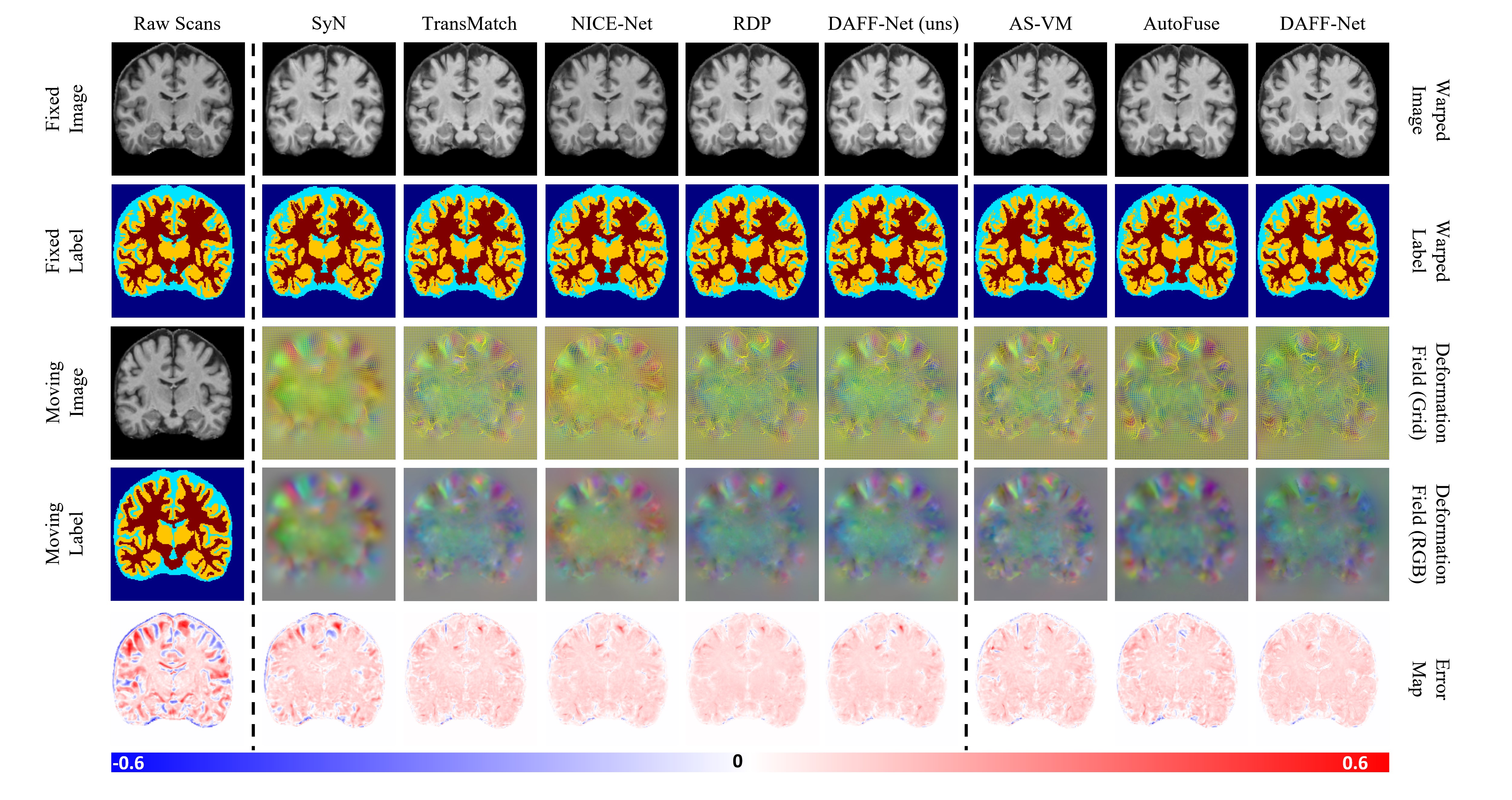}}
	\caption{Visual results of different registration methods on the OASIS3 dataset.}
	\label{fig6}
\end{figure}
\subsection{Evaluation metrics}
We assess the registration accuracy by calculating the Dice score (DSC) \cite{dice1945measures} and the average symmetric surface distance (ASSD) \cite{taha2015metrics} between the fixed segmentation labels $\widetilde{S}_f $ and the warped moving segmentation labels $\widetilde{S}_{m}\circ \phi $. Additionally, the quality of the predicted DDF is evaluated by the percentage of voxels with a non-positive Jacobian determinant (NJD). A better registration should have a higher DSC, lower ASSD, and smaller NJD. All experimental results pass statistical significance testing ($P<0.05$), indicating significant differences between the outcomes of different methods and further validating the effectiveness of the proposed approach.

\begin{figure}[!t]
	\centerline{\includegraphics[width=1.0\textwidth]{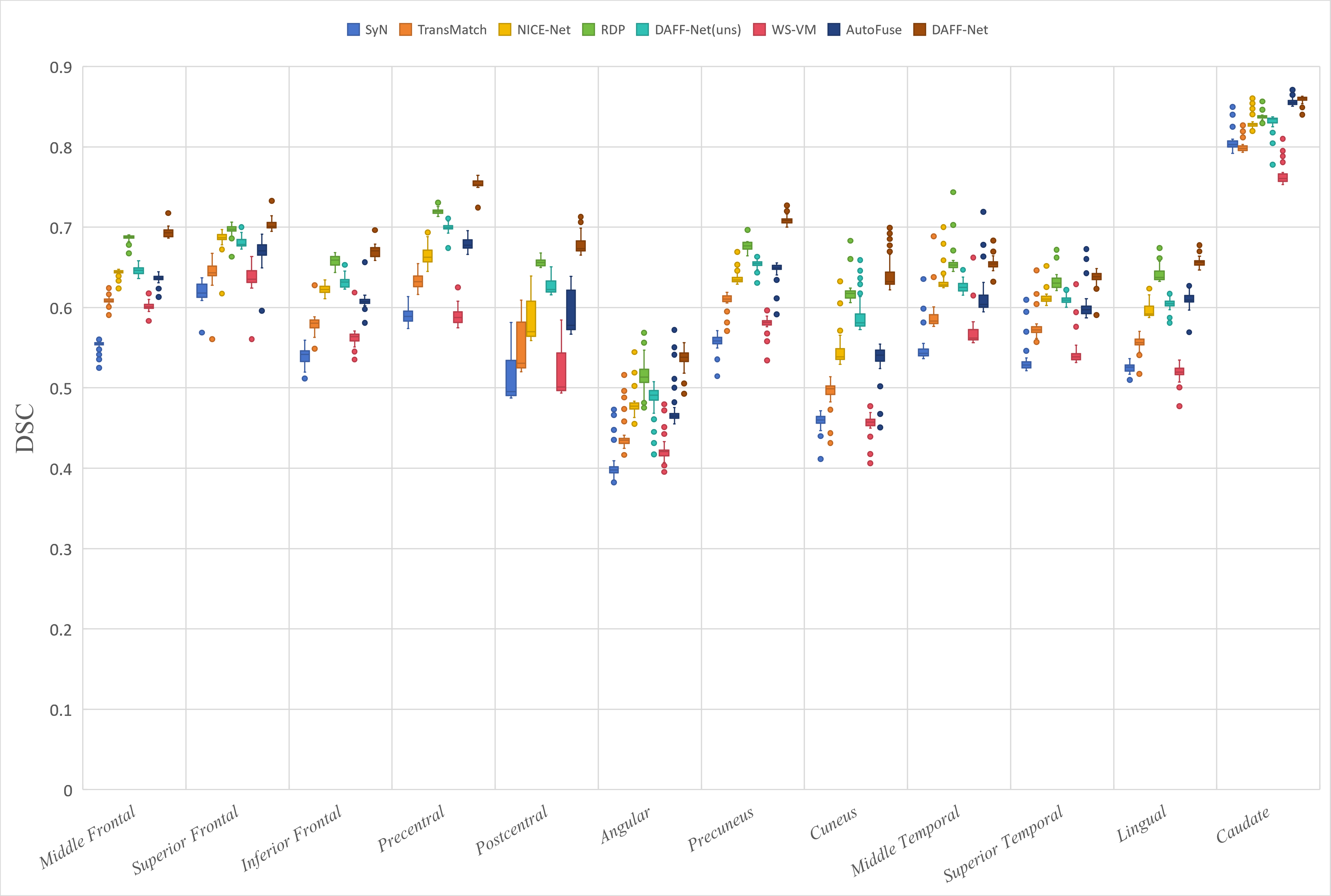}}
	\caption{Box diagrams of DSCs with 12 labels on the OASIS3 dataset.}
	\label{fig7}
\end{figure}

\section{Results}
\subsection{Comparision with other registration methods}
The quantitative evaluation results of different registration methods on the LPBA40, OASIS1, and OASIS3 datasets are reported in Tables \ref{Table 1}, \ref{Table 2} and \ref{Table 3}, respectively.

Across all three datasets, DAFF-Net achieves the best registration accuracy in terms of DSC and ASSD, demonstrating its superior effectiveness in addressing DIR problems. Among unsupervised methods, RDP achieves the highest DSC and ASSD due to its step-by-step recursion strategy, but its DDFs exhibit excessive folding. While our unsupervised variant, DAFF-Net (uns), does not achieve the highest registration accuracy, it still outperformed the other two unsupervised registration methods (TransMatch, NICE-Net) and produces anatomically plausible results with the lowest NJD among unsupervised methods. This indicates that the proposed DAFF module is applicable in the unsupervised version, and DAFF-Net (uns) holds significant potential for future improvements (see section \ref{section 6}).

\begin{table}[!t]
	\centering
	\caption{Quantitative evaluation results of different methods on the LPBA40 dataset. Bolded numbers indicate the best values.}
	\label{Table 1}
	\resizebox{\linewidth}{!}{
		\begin{tabular}{ccccccccccc}
			\hline
			Metric                        & Type & Affine        & SyN                  & TransMatch    & NICE-Net      & RDP           & DAFF-Net (uns) & AS-VM         & AutoFuse      & DAFF-Net               \\ \hline
			\multirow{3}{*}{Reg. DSC(\%) $\uparrow $} & CSF  & 31.98(±4.68)  & 66.43(±2.82)         & 71.59(±2.88)  & 76.43(±2.29)  & 82.04(±1.92)  & 80.50(±2.27)  & 67.90(±3.01)  & 77.96(±2.37)  & \textbf{92.96(±1.22)}  \\
			& GM   & 48.75(±3.67)  & 70.51(±1.90)         & 74.62(±2.46)  & 77.26(±1.61)  & 83.35(±1.28)  & 81.69(±1.51)  & 71.31(±2.60)  & 81.31(±1.59)  & \textbf{92.70(±1.06)}  \\
			& WM   & 61.76(±3.18)  & 80.01(±1.49)         & 84.56(±1.53)  & 85.51(±1.00)  & 90.09(±0.73)  & 88.81(±0.88)  & 81.68(±1.87)  & 88.37(±0.97)  & \textbf{95.17(±0.63)}  \\ \hline
			\multirow{3}{*}{ASSD $\downarrow $}         & CSF  & 1.979(±0.292) & 0.836(±0.058)        & 0.696(±0.051) & 0.603(±0.046) & 0.456(±0.046) & 0.499(±0.050) & 0.778(±0.054) & 0.544(±0.040) & \textbf{0.189(±0.029)} \\
			& GM   & 1.360(±0.182) & 0.734(±0.029)        & 0.658(±0.031) & 0.623(±0.023) & 0.518(±0.023) & 0.548(±0.023) & 0.713(±0.036) & 0.548(±0.022) & \textbf{0.264(±0.028)} \\
			& WM   & 1.847(±0.190) & 0.956(±0.058)        & 0.808(±0.095) & 0.744(±0.041) & 0.525(±0.034) & 0.584(±0.033) & 0.954(±0.119) & 0.594(±0.041) & \textbf{0.266(±0.034)} \\ \hline
			\multirow{3}{*}{Seg. DSC(\%) $\uparrow $} & CSF  & -             & -                    & -             & -             & -             & -             & -             & 97.08(±0.59)  & \textbf{97.76(±0.58)}  \\
			& GM   & -             & -                    & -             & -             & -             & -             & -             & 96.77(±0.58)  & \textbf{97.42(±0.69)}  \\
			& WM   & -             & -                    & -             & -             & -             & -             & -             & 97.90(±0.51)  & \textbf{98.36(±0.46)}  \\ \hline
			NJD(\%) $\downarrow $                       & -    & -             & \textbf{0.01(±0.01)} & 0.63(±0.15)   & 0.05(±0.02)   & 0.44(±0.08)   & 0.02(±0.01)   & 0.65(±0.21)   & 0.06(±0.02)   & 0.14(±0.03)            \\ \hline
		\end{tabular}
	}
\end{table}

\begin{table}[!t]
	\centering
	\caption{Quantitative evaluation results of different methods on the OASIS1 dataset.}
	\label{Table 2}
	\resizebox{\linewidth}{!}{
		\begin{tabular}{ccccccccccc}
			\hline
			Metric                        & Type & Affine        & SyN           & TransMatch    & NICE-Net      & RDP           & DAFF-Net (uns)        & AS-VM         & AutoFuse              & DAFF-Net               \\ \hline
			\multirow{3}{*}{Reg. DSC(\%) $\uparrow $} & CSF  & 49.22(±4.13)  & 65.53(±5.23)  & 75.30(±4.99)  & 75.53(±4.56)  & 78.83(±4.71)  & 77.38(±4.80)         & 74.49(±4.80)  & 82.65(±2.60)          & \textbf{91.87(±1.39)}  \\
			& GM   & 57.81(±3.31)  & 74.13(±2.54)  & 79.80(±2.28)  & 81.00(±2.21)  & 84.14(±2.09)  & 82.61(±2.23)         & 79.18(±2.39)  & 84.78(±1.77)          & \textbf{92.59(±1.32)}  \\
			& WM   & 63.03(±2.27)  & 82.82(±1.29)  & 88.20(±1.10)  & 88.37(±1.10)  & 90.58(±1.09)  & 89.58(±1.07)         & 87.23(±1.32)  & 90.04(±0.84)          & \textbf{94.89(±0.84)}  \\ \hline
			\multirow{3}{*}{ASSD $\downarrow $}         & CSF  & 1.512(±0.118) & 1.034(±0.122) & 0.753(±0.114) & 0.747(±0.107) & 0.652(±0.111) & 0.698(±0.115)        & 0.786(±0.112) & 0.539(±0.050)         & \textbf{0.262(±0.037)} \\
			& GM   & 1.116(±0.050) & 0.753(±0.044) & 0.628(±0.047) & 0.610(±0.044) & 0.544(±0.046) & 0.577(±0.048)        & 0.647(±0.046) & 0.516(±0.030)         & \textbf{0.287(±0.030)} \\
			& WM   & 1.704(±0.087) & 0.891(±0.060) & 0.661(±0.061) & 0.651(±0.062) & 0.531(±0.063) & 0.584(±0.061)        & 0.713(±0.072) & 0.545(±0.049)         & \textbf{0.305(±0.054)} \\ \hline
			\multirow{3}{*}{Seg. DSC(\%) $\uparrow $} & CSF  & -             & -             & -             & -             & -             & -                    & -             & \textbf{97.50(±0.58)} & 96.74(±0.64)           \\
			& GM   & -             & -             & -             & -             & -             & -                    & -             & \textbf{97.15(±0.83)} & 96.42(±0.83)           \\
			& WM   & -             & -             & -             & -             & -             & -                    & -             & \textbf{97.86(±0.89)} & 97.33(±0.78)           \\ \hline
			NJD(\%) $\downarrow $                      & -    & -             & 0.04(±0.02)   & 1.11(0.09)    & 0.06(±0.01)   & 0.89(0.07)    & \textbf{0.03(±0.00)} & 1.19(±0.11)   & 0.06(±0.01)           & 0.17(±0.02)            \\ \hline
		\end{tabular}
	}
\end{table}

Among all methods that improve registration performance using segmentation, the approach using auxiliary segmentation labels to provide anatomical information for registration (AS-VM) yields unsatisfactory results. Compared to AutoFuse, the proposed DAFF-Net achieves higher registration accuracy across all three datasets. Although the fold on the DDFs is slightly higher, DAFF-Net still produces anatomically plausible results. As a deep learning framework for joint registration and segmentation, DAFF-Net produces a slightly lower DSC on the OASIS1 and OASIS3 datasets in terms of segmentation accuracy from the segmentation network. This can be interpreted as the design of our network architecture focusing on segmentation as an auxiliary task to provide structural information for registration. Therefore, the primary goal of DAFF-Net is to achieve more precise and plausible registration.

Figures \ref{fig4}, \ref{fig5} and \ref{fig6} present registration result slices for various methods on the LPBA40, OASIS1, and OASIS3 datasets, respectively. Compared to other methods, DAFF-Net exhibits higher precision in the registered images, validating our approach of fully leveraging the correlation between registration and segmentation within a joint learning framework, thereby maximizing the benefit of segmentation in enhancing registration. In the last row of each set of images, we visualize the voxel-wise difference maps (error maps) between the warped moving and fixed images under each method. Although the error maps of our unsupervised variant are not the best among the unsupervised methods, they still outperform TransMatch and NICE-Net. Our complete DAFF-Net shows the best error maps among three leveraging segmentation-enhance registration methods (last three columns). The results from the third and fourth rows clearly show that both DAFF-Net and its unsupervised variant achieve plausible smooth DDFs.

\begin{table}[]
		\centering
	\caption{Quantitative evaluation results of different methods on the OASIS3 dataset.}
	\label{Table 3}
	\resizebox{\linewidth}{!}{
	\begin{tabular}{ccccccccccc}
		\hline
		Metric                        & Type & Affine        & SyN                  & TransMatch    & NICE-Net      & RDP           & DAFF-Net (uns) & AS-VM         & AutoFuse              & DAFF-Net               \\ \hline
		\multirow{3}{*}{Reg. DSC(\%) $\uparrow $} & CSF  & 52.72(±3.56)  & 74.97(±2.80)         & 81.46(±1.90)  & 81.55(±1.92)  & 82.76(±1.89)  & 82.49(±1.84)  & 79.84(±2.23)  & 85.26(±1.08)          & \textbf{87.20(±1.07)}  \\
		& GM   & 54.04(±2.17)  & 72.05(±1.83)         & 77.55(±1.48)  & 77.68(±1.54)  & 79.81(±1.60)  & 78.89(±1.56)  & 75.92(±1.74)  & 81.85(±1.04)          & \textbf{83.74(±1.08)}  \\
		& WM   & 61.13(±2.98)  & 79.56(±1.72)         & 85.14(±1.27)  & 84.79(±1.36)  & 86.41(±1.36)  & 85.60(±1.38)  & 84.18(±1.44)  & 87.68(±0.94)          & \textbf{89.12(±0.99)}  \\ \hline
		\multirow{3}{*}{ASSD $\downarrow $}         & CSF  & 1.583(±0.098) & 0.846(±0.059)        & 0.627(±0.026) & 0.621(±0.028) & 0.572(±0.027) & 0.590(±0.026) & 0.670(±0.034) & 0.507(±0.017)         & \textbf{0.455(±0.018)} \\
		& GM   & 1.050(±0.051) & 0.714(±0.036)        & 0.617(±0.026) & 0.616(±0.027) & 0.590(±0.027) & 0.601(±0.026) & 0.646(±0.031) & 0.526(±0.017)         & \textbf{0.504(±0.020)} \\
		& WM   & 1.571(±0.077) & 0.877(±0.048)        & 0.682(±0.035) & 0.690(±0.036) & 0.623(±0.037) & 0.658(±0.037) & 0.731(±0.042) & 0.559(±0.024)         & \textbf{0.524(±0.029)} \\ \hline
		\multirow{3}{*}{Seg. DSC(\%) $\uparrow $} & CSF  & -             & -                    & -             & -             & -             & -             & -             & \textbf{93.61(±0.55)} & 90.45(±0.87)           \\
		& GM   & -             & -                    & -             & -             & -             & -             & -             & \textbf{91.89(±0.78)} & 87.77(±0.83)           \\
		& WM   & -             & -                    & -             & -             & -             & -             & -             & \textbf{94.66(±0.63)} & 91.62(±0.81)           \\ \hline
		NJD(\%) $\downarrow $                       & -    & -             & \textbf{0.02(±0.01)} & 1.20(±0.12)   & 0.05(±0.01)   & 1.03(±0.10)   & 0.03(±0.00)   & 1.42(±0.17)   & 0.04(±0.01)           & 0.10(±0.01)            \\ \hline
	\end{tabular}
}
\end{table}

The distributions of DSC of specific anatomical regions on the OASIS3 dataset are illustrated in Figure \ref{fig7}. We show the DSC distributions of twelve anatomical labels, including middle frontal, superior frontal, inferior frontal, precentral, postcentral, angular, precuneus, cuneus, middle temporal, superior temporal, lingual, and caudate. It can be observed that the proposed DAFF-Net achieves the highest DSC in all these regions, indicating its excellent performance in handling areas with large and complex deformations. Additionally, the unsupervised variant DAFF-Net (uns) also performs well in these regions. Compared to NICE-Net, DAFF-Net (uns) achieves higher DSC in most regions, further demonstrating that the DAFF module effectively captures complex information in images by fusing feature information.

\subsection{Ablation study} \label{section 5.2}

The novelty of our proposed DAFF-Net is to maximize the correlation between different tasks by utilizing supervised segmentation to effectively learn the anatomical structure of ROIs. Additionally, we utilize the proposed DAFF module to process feature information from various tasks, capturing the most beneficial feature maps for the registration task during the feature fusion process. To validate our design motivation, we perform the ablation experiments, with Figure \ref{fig3} illustrating the structure of each ablation network.

The encoder and decoder architecture of PyramidReg (Figure \ref{fig3} (a)) is identical to that of DAFF-Net, utilizing the same conv block and flow estimator block (FEB). We select it as our baseline due to its relatively strong performance in handling DIR problems. In AuxReg (Figure \ref{fig3} (b)), segmentation labels (Seg. Aux.) are used to provide auxiliary information for registration, allowing us to assess whether this approach improves the performance of PyramidReg. In SimSReg (Figure \ref{fig3} (c)), we introduce a simple U-Net-based segmentation network as an auxiliary task (Mul. Task) where the two task networks are separated. Next, in GloSReg (Figure \ref{fig3} (d)), we consider the coupling (Coup.) between registration and segmentation, employing a global encoder to extract common features and designing task-specific decoders to complete the registration and segmentation tasks individually. To evaluate whether simple concatenation can effectively achieve feature fusion (Feat. Fus.) between the two tasks, CcSReg (Figure \ref{fig3} (e)) concatenate the feature maps of these tasks at different scales and feds them into a convolutional layer without DAFF module. Finally, DAFF-Net (Figure \ref{fig3} (f)) is our complete version by replacing simple concatenation of CcSReg with DAFF modules, whose objective is to capture the most beneficial information for registration from the fusion features.

\begin{table}[]
	\centering
	\caption{Ablation study of individual components in our proposed DAFF-Net on the LPBA40 dataset. }
	\label{Table 4}
	\resizebox{\linewidth}{!}{
	\begin{tabular}{cccccccccc}
		\hline
		Method       & Seg. Aux. & Mul. Task & Coup.     & Feat. Fus. & DAFF    & Reg. Dice(\%) $\uparrow $         & Seg. Dice(\%) $\uparrow $         & ASSD $\downarrow $                   & NJD(\%) $\downarrow $              \\ \hline
		PyramidReg   &           &           & \textbf{} &            &         & 82.02(±4.19)          & -                     & 0.591(±0.028)          & \textbf{0.02(±0.01)} \\
		AuxReg      & $\surd$   &           &           &            &         & 82.30(±4.11)          & -                     & 0.583(±0.027)          & \textbf{0.02(±0.01)} \\
		SimSReg      & $\surd$   & $\surd$   &           &            &         & 89.97(±2.37)          & 97.27(±0.76)          & 0.364(±0.027)          & 0.14(±0.03)          \\
		GloSReg      & $\surd$   & $\surd$   & $\surd$   &            &         & 90.34(±2.01)          & 97.18(±0.74)          & 0.355(±0.028)          & 0.13(±0.03)          \\
		CcSReg & $\surd$   & $\surd$   & $\surd$   & $\surd$    &         & 91.50(±1.89)          & 97.42(±0.74)          & 0.312(±0.027)          & 0.14(±0.03)          \\
		DAFF-Net     & $\surd$   & $\surd$   & $\surd$   & $\surd$    & $\surd$ & \textbf{93.61(±1.49)} & \textbf{97.85(±0.70)} & \textbf{0.240(±0.028)} & 0.14(±0.03)          \\ \hline
	\end{tabular}
}
\end{table}

To train PyramidReg, we use the conventional registration loss function $L_{reg}$ (consists of $L_{sim}$, $L_{smooth}$ and $L_{njd}$) and apply the same hyperparameter $\lambda_{0}$ as mentioned in Section \ref{section 4.3}. For AuxReg, in addition to the conventional registration loss $L_{reg}$, we also calculate the Dice loss and Focal loss ($L_{FocalDice}$) between the fixed segmentation labels ($\widetilde{S}_f$) and the warped moving segmentation labels ($\widetilde{S}_{m}\circ \phi$) during model training (the weight of $L_{FocalDicee}$ consistent with $\lambda_{2}$ as described in Section \ref{section 4.3}). Finally, SimSReg, GloSReg, and DAFF-Net(CC) are trained using the complete loss function $L_{total}$ from Eq. \ref{eq11}, with the same parameter settings as DAFF-Net.

\begin{figure}[!t]
	\centerline{\includegraphics[width=1.0\textwidth]{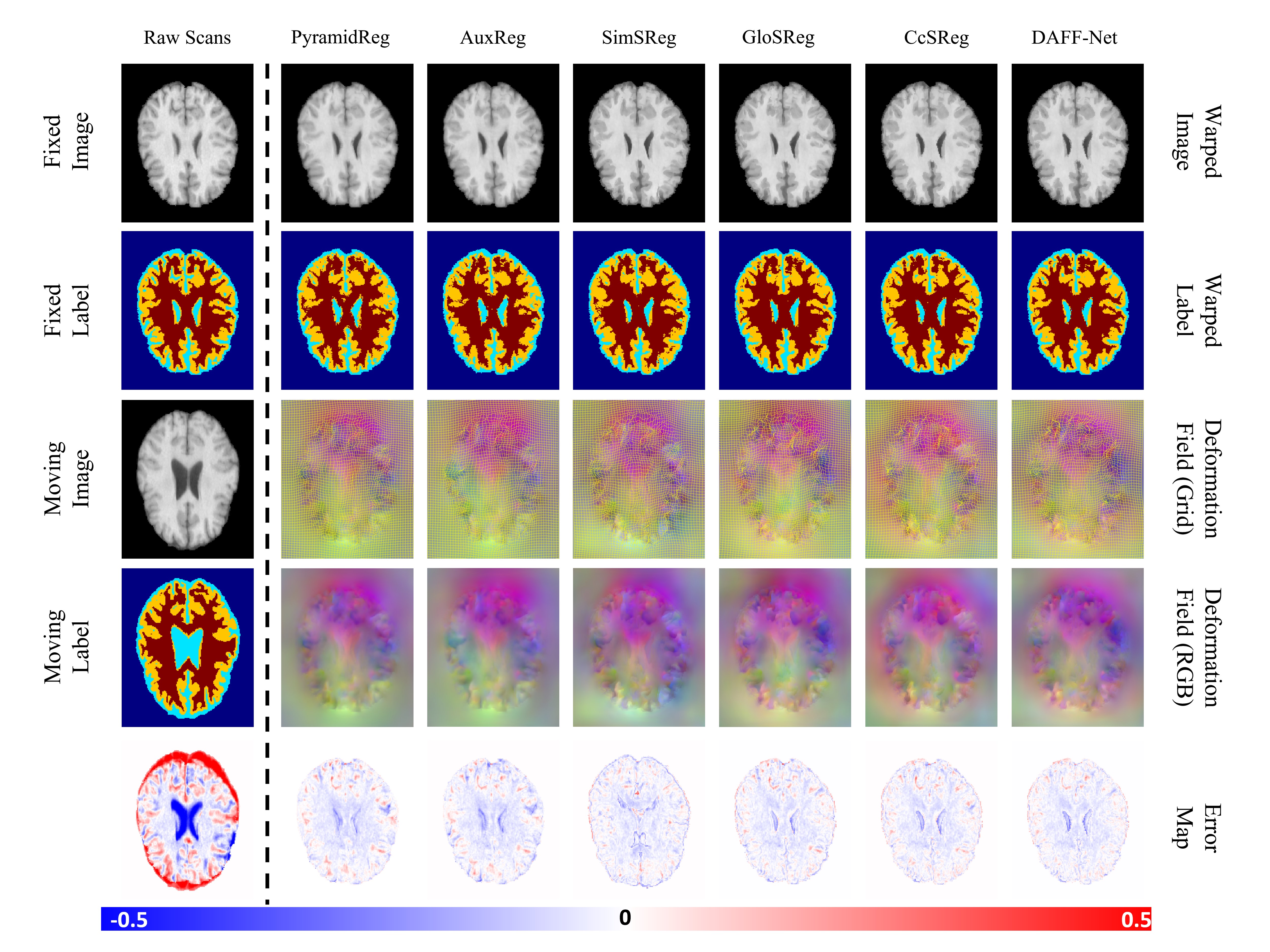}}
	\caption{Visual results of different network variants on the LPBA40 dataset.}
	\label{fig8}
\end{figure}

Figure \ref{fig8} shows the registration result slices of different ablation network variants on the LPBA40 dataset. It can be observed that the proposed DAFF-Net achieves the smallest error between the warped moving images and the fixed images. On the other hand, while the variants that include a segmentation network as an auxiliary task do not perform as well as PyramidReg and AuxReg in terms of DDFs smoothness, their overall performance remains plausible compared to TransMatch, RDP, and AS-VM. 

The quantitative evaluation results of different ablation network variants are presented in Table \ref{Table 4}. Compared to PyramidReg, using segmentation labels to provide auxiliary information for registration (AuxReg) only results in a minor improvement in registration performance. This may be because the selected segmentation labels only cover three brain tissues, resulting in sparse gradient information when simply calculating the loss between the fixed segmentation labels and the warped moving segmentation labels. This sparsity makes the network less sensitive to small deformations during optimization, preventing it from effectively capturing anatomical details of the ROIs, which in turn limits the improvement in registration accuracy. After incorporating segmentation as an auxiliary task, SimSReg shows a significant improvement in registration accuracy compared to AuxReg, indicating that multi-task learning can better leverage the correlation between registration and segmentation. Additionally, the comprehensive fusion loss $L_{fuse}$ further enhances the alignment accuracy in the ROIs. By using a global encoder to extract common features for both registration and segmentation, GloSReg achieves slightly higher registration accuracy than SimSReg, demonstrating that this architecture can further enhance the coupling between the two tasks. CcSReg improves registration accuracy by over $1\%$ compared to GloSReg, demonstrating that feature fusion can further enhance the correlation between tasks in multi-task learning. Finally, when it comes to feature fusion, the complete network (DAFF-Net) using our proposed DAFF module achieves over a $2\%$ improvement in registration accuracy compared to a network using simple concatenation (CcSReg). This demonstrates that by leveraging both global and local attentions to process feature maps, the DAFF module effectively decouples fusion features, focusing on those beneficial for the registration task. Additionally, by handling features from both high-frequency and low-frequency perspectives, the DAFF module enhances the network's ability to capture complex and comprehensive information in the images. The findings indicate that each of the proposed components makes a substantial contribution.

\subsection{Analysis of parameter setting} \label{section 5.3}

\begin{table}[]
	\centering
	\caption{Quantitative evaluation results on LPBA40 data set under different $\lambda_{1}$ and $\lambda_{2}$ parameters training. }
	\label{Table 5}
	\resizebox{\linewidth}{!}{
	\begin{tabular}{ccccccccc}
		\hline
		Metric                        & Type & $\lambda_{1}0.5$/$\lambda_{2}1$       & $\lambda_{1}1$/$\lambda_{2}1$                & $\lambda_{1}5$/$\lambda_{2}1$                & $\lambda_{1}10$/$\lambda_{2}1$        & $\lambda_{1}1$/$\lambda_{2}0.5$               & $\lambda_{1}1$/$\lambda_{2}5$                  & $\lambda_{1}1$/$\lambda_{2}10$                 \\ \hline
		\multirow{3}{*}{Reg. DSC(\%)} & CSF  & 92.96(±1.22)  & 92.54(±1.18)         & 92.65(±1.34)         & 92.17(±1.17)  & 91.27(±1.36)          & 94.24(±0.96)           & \textbf{94.29(±0.95)}  \\
		& GM   & 92.70(±1.06)  & 92.36(±1.05)         & 92.22(±1.22)         & 92.31(±0.95)  & 91.38(±1.05)          & 93.99(±0.86)           & \textbf{94.02(±0.90)}  \\
		& WM   & 95.17(±0.63)  & 95.04(±0.69)         & 94.68(±0.73)         & 95.09(±0.53)  & 94.53(±0.60)          & 96.13(±0.50)           & \textbf{96.17(±0.59)}  \\ \hline
		\multirow{3}{*}{ASSD}         & CSF  & 0.189(±0.029) & 0.201(±0.028)        & 0.199(±0.032)        & 0.211(±0.029) & 0.235(±0.031)         & \textbf{0.157(±0.023)} & \textbf{0.157(±0.024)} \\
		& GM   & 0.264(±0.028) & 0.273(±0.030)        & 0.276(±0.030)        & 0.274(±0.025) & 0.300(±0.025)         & 0.225(±0.024)          & \textbf{0.223(±0.026)} \\
		& WM   & 0.266(±0.034) & 0.274(±0.038)        & 0.291(±0.038)        & 0.271(±0.029) & 0.298(±0.031)         & \textbf{0.219(±0.028)} & 0.221(±0.035)          \\ \hline
		\multirow{3}{*}{Seg. DSC(\%)} & CSF  & 97.76(±0.58)  & 97.79(±0.64)         & 97.99(±0.71)         & 98.00(±0.67)  & \textbf{98.09(±0.43)} & 97.40(±0.53)           & 96.68(±0.60)           \\
		& GM   & 97.42(±0.69)  & 97.33(±0.89)         & 97.38(±1.03)         & 97.59(±0.65)  & \textbf{97.69(±0.57)} & 97.07(±0.64)           & 96.38(±0.71)           \\
		& WM   & 98.36(±0.46)  & 98.21(±0.69)         & 98.17(±0.77)         & 98.41(±0.39)  & \textbf{98.46(±0.42)} & 98.14(±0.41)           & 97.75(±0.52)           \\ \hline
		NJD(\%)                       & -    & 0.14(±0.03)   & \textbf{0.13(±0.03)} & \textbf{0.13(±0.03)} & 0.14(±0.03)   & 0.08(±0.02)           & 0.51(±0.08)            & 0.82(±0.10)            \\ \hline
	\end{tabular}
}
\end{table}

Table \ref{Table 5} tabulates the quantitative results for a range of $\lambda_{1}$ and $\lambda_{2}$ settings. Due to GPU limitations, we fix one hyperparameter at 1 while varying the other between [0.5, 1, 5, 10]. As $\lambda_{1}$ (the weight of $L_{seg}$) keeps increasing, although the segmentation DSC improves, the registration accuracy decreases, which contradicts our primary goal of optimizing the registration task. Furthermore, as $\lambda_{2}$ (the weight of $L_{fusion}$) keep increasing, the registration DSC improves, but the DDFs become overly folded when $\lambda_{2}$ is set to 5 or 10. Therefore, considering both registration accuracy and DDFs smoothness, setting $\lambda_{1}$ to 0.5 and $\lambda_{2}$ to 1 provides a good balance for weakly supervised anatomical constraints in the registration task.

\section{Conclusions}  \label{section 6}
In this paper, we propose a novel multi-task learning framework based on multi-scale dual attention frequency fusion for deformable medical image registration, which leverages an auxiliary segmentation task to provide anatomical information for registration, effectively exploiting the correlation between registration and segmentation tasks. The proposed DAFF-Net not only efficiently transfers anatomical prior knowledge from segmentation to registration but also introduces a dual attention frequency fusion (DAFF) module to finely process the features of both tasks at different scales. DAFF-Net not only improves registration accuracy but also ensures the plausibleness of the registration results.

Leveraging the DAFF module's exceptional ability to process fusion features, we extend it to unsupervised registration methods (DAFF-Net (uns)). Compared to current unsupervised registration methods, DAFF-Net (uns) demonstrates superior registration performance while maintaining smooth deformation. However, due to the lack of constraint on the auxiliary decoder in the unsupervised variant, DAFF-Net (uns) still has significant room for improvement. Our future work will focus on exploring effective constraints for the auxiliary decoder during unsupervised training, enabling it to provide valuable feature information for registration. This will further enhance the DAFF module's ability to capture the most beneficial features.

\section*{CRediT authorship contribution statement}
\textbf{Hongchao Zhou}: Conceptualization, Data curation, Validation, Visualization, Writing-original draft, Writing-review \& editing. \textbf{Shunbo Hu}: Conceptualization, Writing-review \& editing, Funding acquisition, Project administration, Resources, Supervision. 

\section*{Declaration of competing interest}
The authors declare that they have no known competing financial interests or personal relationships that could have appeared to influence the work reported in this paper.

\section*{Code availability}
The code used to train and evaluate the modules is available from the corresponding authors upon request.

\section*{Acknowledgments}
This work was supported by project ZR2023MF062 supported by Shandong Provincial Natural Science Foundation.

\bibliography{ref.bib}
\bibliographystyle{elsarticle-num}

\end{document}